\documentclass{article}
\usepackage{colm2024_conference}
\usepackage{amsmath,amssymb}
\usepackage{graphicx}
\usepackage{xcolor}
\usepackage{fontawesome5}
\definecolor{githubblue}{RGB}{0,0,128}
\fancypagestyle{firstpage}{%
  \fancyhead{}%
  \lhead{\includegraphics[height=10pt]{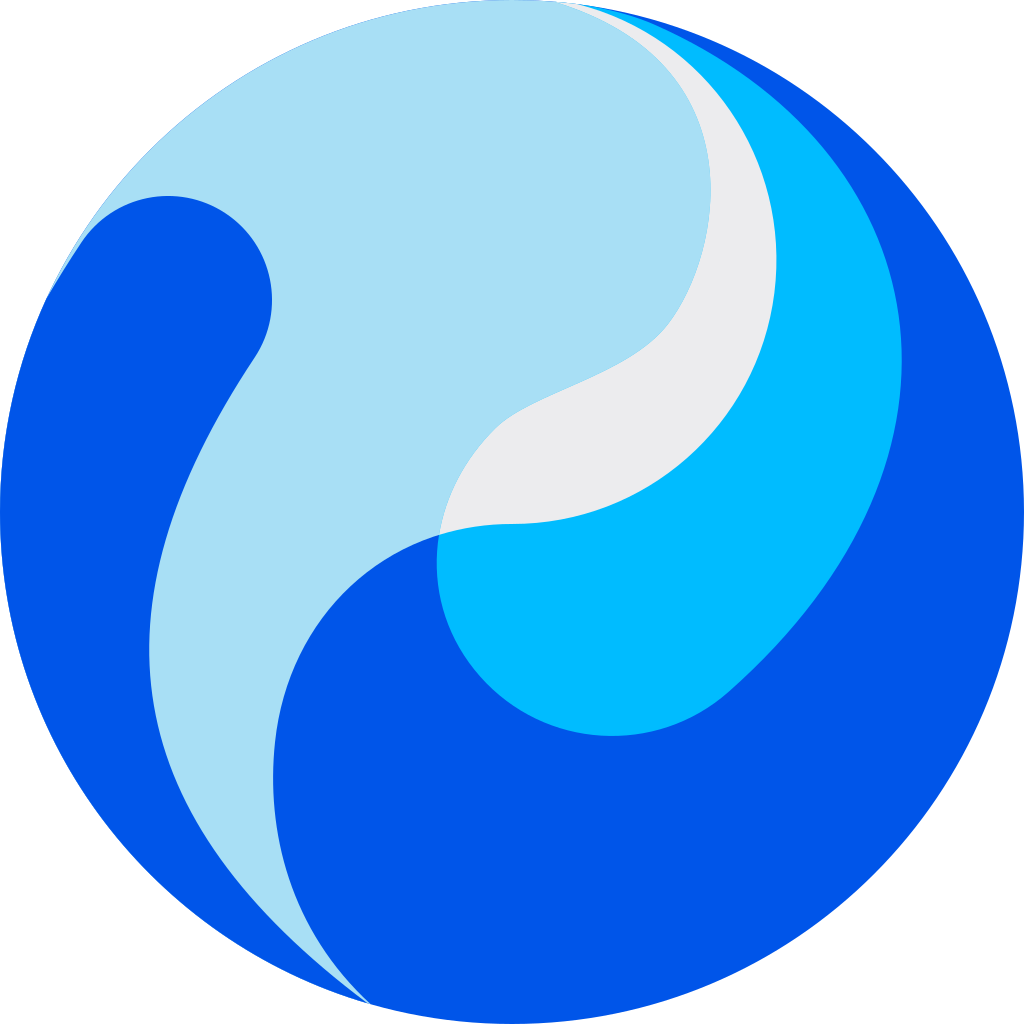}\ Tencent HY}%
  \renewcommand{\headrulewidth}{1pt}%
}
\usepackage{booktabs,array,multirow}
\usepackage{colortbl}
\usepackage{microtype}
\usepackage{enumitem}
\usepackage{wrapfig}
\usepackage{tocloft}

\cftsetindents{section}{0em}{1.5em}
\cftsetindents{subsection}{1.5em}{2.3em}
\cftsetindents{subsubsection}{3.8em}{3.2em}
\hypersetup{hidelinks,pdftitle={Does Native 3D Texture Generation Necessarily Require 3D Assets for Training?},pdfauthor={Jiangshan Wang, Zeqiang Lai, Jiayi Guo, Xin Yang, Xin Huang, Jiarui Chen, Ziheng Ouyang, Chunchao Guo, Xiangyu Yue}}

\title{Does Native 3D Texture Generation Necessarily Require 3D Assets for Training?}

\author{{\href{https://github.com/wangjiangshan0725/Tex-Zero}{\resizebox{0.62\textwidth}{!}{\textcolor{black}{\faGithub}\hspace{0.5em}\textcolor{githubblue}{\texttt{https://github.com/wangjiangshan0725/Tex-Zero}}}}}\\[0.85em]
\textbf{Jiangshan Wang}$^{1,2}$ \quad
\textbf{Zeqiang Lai}$^{1,2\dagger}$ \quad
\textbf{Jiayi Guo}$^{3}$ \\
\textbf{Xin Yang}$^{2}$ \quad
\textbf{Xin Huang}$^{2}$ \quad
\textbf{Jiarui Chen}$^{2,4,5}$ \quad
\textbf{Ziheng Ouyang}$^{2,6}$ \\
\textbf{Chunchao Guo}$^{2*}$ \quad
\textbf{Xiangyu Yue}$^{1*}$ \\
$^{1}$MMLab, CUHK \quad
$^{2}$Tencent Hunyuan \quad
$^{3}$Tsinghua University \\
$^{4}$Fudan University \quad
$^{5}$Shanghai Innovation Institute \quad
$^{6}$Nankai University \\
}

\newcommand{\G}{\mathcal{G}}
\newcommand{\C}{\mathcal{C}}
\newcommand{\I}{\mathcal{I}}

\newcommand{\R}{\mathbb{R}}

\begin{document}
\pagestyle{fancy}
\maketitle
\pagestyle{fancy}\fancyhead{}
\fancyhead[L]{\includegraphics[height=10pt]{hunyuan-logo.png}\ Tencent HY}
\renewcommand{\headrulewidth}{1pt}

\renewcommand{\thefootnote}{}
\footnotetext{\textsuperscript{$\dagger$} Project lead. \textsuperscript{$*$} Corresponding authors.}


\begin{abstract}
Native 3D texture generation synthesizes colors directly in 3D space for a given geometry, conditioned on multi-view reference images. It is generally believed that training such models requires large-scale, high-quality real 3D asset data, whose acquisition remains a long-standing and challenging problem. In this work, we propose \textbf{Tex-Zero}, demonstrating that a high-fidelity native 3D texture generation framework can be trained without 3D assets. Our key observation is that only high-quality and fine-grained color information is essential for 3D texture training, while the required geometric information is less critical and can be manually constructed rather than obtained from real 3D assets. This finding makes it possible to transform abundant, high-quality 2D images into effective training samples for 3D texture generation.
Specifically, we convert high-quality 2D images into 3D training samples by representing each image as a plane in 3D space and applying patch-wise random rotations and aggregation to construct complex geometric structures. Using these constructed image data, we train the Tex-Zero VAE, which can reconstruct real 3D assets with high quality despite never observing them during training. 
Building upon the Tex-Zero VAE, we train the Tex-Zero DiT also exclusively on the constructed image data, where the conditioning 2D multi-view images are transformed into planes in 3D space and also encoded by the Tex-Zero VAE, thereby reducing the representation gap and improving generation quality. Extensive experiments show that {Tex-Zero} generates high-fidelity 3D textures with fine-grained details solely using images as training data, offering a promising perspective on the data paradigm for scaling 3D texture generation.

\end{abstract}
\begin{figure}[h]
    \centering
    \includegraphics[width=\textwidth]{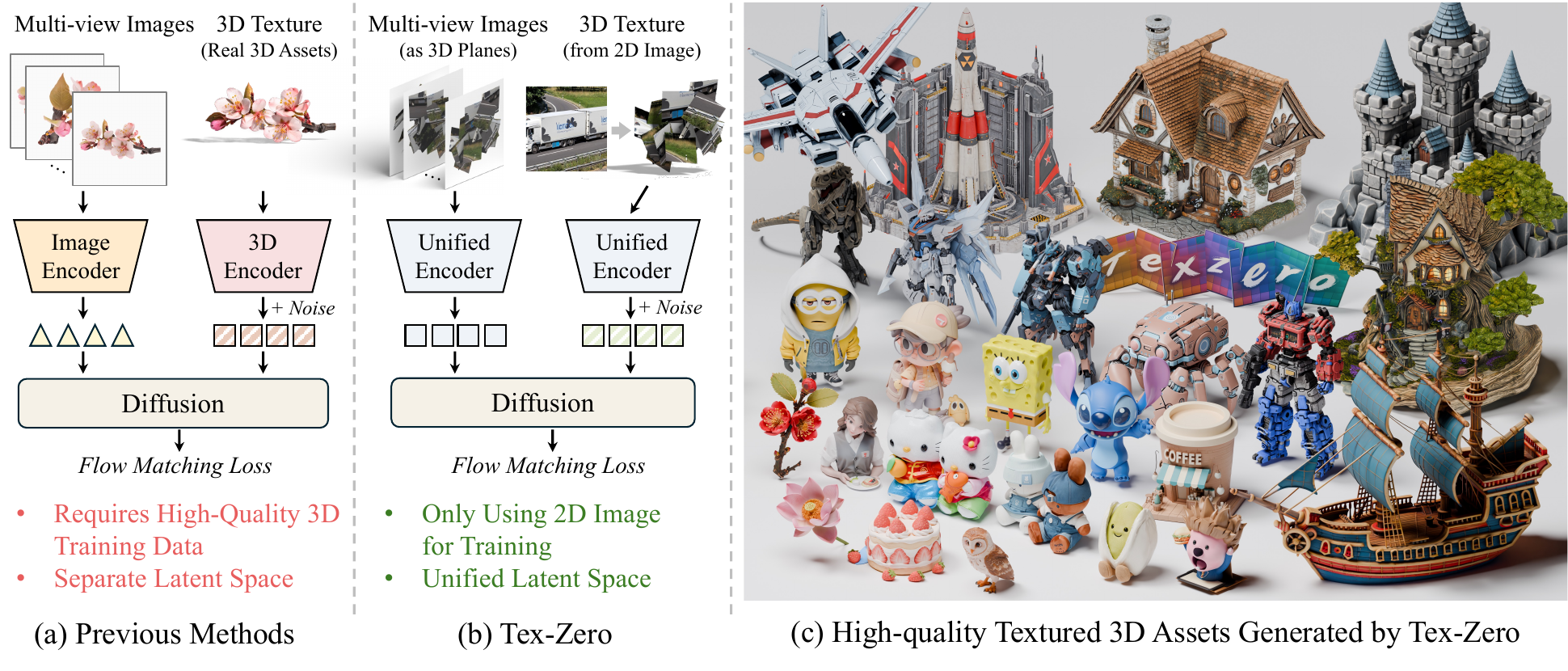}
    \caption{Compared with previous methods, Tex-Zero trains the VAE and DiT without any real 3D assets and learns a unified latent space for 2D images and 3D textures. With only images as the training data, Tex-Zero achieves high-fidelity generation of real 3D assets.}
    \label{fig:teaser}
\end{figure}

\section{Introduction}
\label{sec:intro}

Texture generation aims to generate detailed textures for 3D objects from reference multi-view images while preserving geometric alignment and global consistency. In recent years, native 3D texture generation has emerged as a promising paradigm that predicts colors directly in 3D space \citep{he2026hitem3d,lai2025natex}. Unlike conventional methods that construct 3D textures by the reprojection process from 2D views \citep{richardson2023texture,chen2023text2tex,zeng2024paint3d,liu2024syncmvd,yeh2024texturedreamer,huo2024texgen}, native texture generation avoids errors introduced by projection and fusion, providing a natural formulation for geometry-aligned and globally coherent texturing. 

Training such models typically requires high-quality 3D assets, where a large number of spatial points and their corresponding colors collectively define a semantically meaningful 3D object with complex geometry and rich texture details. Unlike natural language and 2D image or video data, high-quality 3D asset data are scarce and often require specialized equipment and costly data acquisition processes, whose acquisition remains a long-standing and challenging problem.
Seeking an alternative to costly 3D texture data, we notice that 2D images and 3D textures share a common data structure: both assign colors to spatial locations (In an image, each color is assigned to a pixel in 2D space, while in 3D textured data, each color is assigned to a point on a 3D surface). Unlike 3D texture data, large-scale high-quality 2D images are readily available. This observation naturally leads to a question: Can we assign the rich color information in abundant, high-quality 2D images to points in 3D space, thereby turning 2D images into effective training data for 3D texture generation?

In this work, we provide an affirmative answer to this question by proposing \textbf{Tex-Zero}, 
in which we find that 3D texture generation can be trained \textit{exclusively} on 2D image data. Our key finding is that effective training samples for native 3D texture generation require high-quality, fine-grained texture details, while the geometric information is less critical and can be manually constructed without relying on semantically meaningful 3D assets.
Specifically, we develop a straightforward data construction pipeline that converts 2D images into effective 3D training samples for native 3D texture generation. We first treat each 2D image as a planar surface in 3D space, thereby representing the image in the form of 3D data. 
Then, we apply random patch-wise rotations in 3D space and aggregate the patches to reduce floaters. This process introduces complex local geometric structures and occlusion patterns while preserving the detailed texture information of the original image within each patch. We find that such samples are sufficient to effectively train both the 3D VAE and DiT.

Solely using data constructed from images, we first train a 3D VAE  (i.e., \textbf{Tex-Zero VAE}). Despite never seeing real 3D assets during training, the Tex-Zero VAE can faithfully reconstruct real 3D assets at inference time. Moreover, since our training data are derived from high-quality 2D images, we find that the Tex-Zero VAE can also effectively reconstruct 2D images when they are represented as planes in 3D space, preserving fine-grained visual details. In contrast, existing 3D VAEs trained on 3D textured data struggle to achieve high-fidelity reconstruction of 2D images and often produce blurry and low-quality reconstructions. These results imply that our data construction enables a unified 2D-3D VAE, providing a high-quality shared latent space for 2D images and 3D textures.

Building upon the Tex-Zero VAE, we further train the \textbf{Tex-Zero DiT} using only image data. Leveraging the strong reconstruction capability of the Tex-Zero VAE, we adopt a design inspired by existing image editing pipelines \citep{wu2025qwen}, where the input 2D multi-view images are transformed into planes in 3D space and also encoded by the Tex-Zero VAE, then injected into Tex-Zero DiT as conditions. This design allows the multi-view conditions and the target 3D texture to be represented in the unified latent space, facilitating more effective training and better generation performance. Extensive experiments demonstrate that Tex-Zero effectively transfers the knowledge learned on constructed data to real 3D assets at inference time, achieving high-fidelity 3D texture generation and outperforming representative baselines trained on real textured 3D data. Unleashing the potential of large-scale 2D data for native 3D texture generation, Tex-Zero offers a promising perspective on overcoming the 3D data scarcity bottleneck. Furthermore, we believe that it could suggest broader possibilities for sourcing and constructing effective training data beyond real 3D assets, potentially taking a step toward scaling 3D texture generation. 

Our main contributions are summarized as follows:
\begin{itemize}[leftmargin=*, itemsep=1pt, topsep=2pt, parsep=0pt]
    \item We are the first to propose a data construction pipeline that transforms large-scale 2D images into effective training data for native 3D texture generation, revealing that the geometric structures of training data need not be semantically meaningful.

    \item We represent conditioning multi-view images as planes in 3D space and encode them together with target 3D textures using the Tex-Zero VAE to eliminate the representation gap, which is enabled by the strong encoding capability of Tex-Zero VAE learned from image-derived data.

    \item We demonstrate that {Tex-Zero} can generate high-fidelity 3D textures with fine-grained details, suggesting a promising direction for scaling 3D generative models through abundant image data. 
\end{itemize}

\section{Related Works}
\label{sec:related}

\textbf{3D Texture Generation.}
3D texture generation aims to synthesize coherent textures for a given 3D geometry. Existing methods commonly synthesize multi-view images as conditions for generation robustness and visual quality. Conventional view-based methods construct textures by projecting, fusing, and baking these images onto the target geometry \citep{zeng2024paint3d,huo2024texgen,cheng2025mvpaint,he2025materialmvp,hunyuan2025pbr}. However, such multi-stage pipelines are susceptible to accumulated projection errors and cross-view inconsistencies, compromising texture fidelity and coherence. In contrast, native methods generate textures directly in geometry-aligned representations, including UV maps \citep{yu2024texgen}, octree-aligned 3D Gaussians \citep{xiong2025texgaussian}, continuous texture functions \citep{liang2025unitex}, native surface colors \citep{lai2025natex,he2026hitem3d}, and structured geometry–appearance latents \citep{xiang2025trellis,xiang2025trellis2}. Despite their improved geometric consistency, these methods rely heavily on high-quality 3D assets for training.

\textbf{Learning 3D models from synthetic data.}
Collecting high-quality 3D data is costly and time-consuming, motivating the use of synthetic data to reduce the reliance of 3D learning on manually collected assets. LRM-Zero trains reconstruction models on procedurally generated textured shapes, while MegaSynth extends this strategy to large-scale scene reconstruction \citep{xie2024lrmzero,jiang2025megasynth}. VFusion3D instead employs a video diffusion model to generate millions of synthetic multi-view examples for training a feed-forward 3D reconstruction model \citep{han2024vfusion3d}. These studies demonstrate that carefully designed synthetic data can generalize to real-world inputs without fully matching the semantic distribution of real data. Nevertheless, leveraging synthetic data to train native 3D texture generation models remains largely unexplored.

\section{Methods}
\label{sec:method}

Tex-Zero is a high-fidelity native texture generation framework trained solely on 2D images. In this section, we first describe how 2D images are transformed for 3D texture training. We then present the training and inference procedures of the Tex-Zero VAE, which constructs a shared latent space for both 2D images and 3D assets. Finally, we introduce the Tex-Zero DiT and demonstrate that, despite being trained without any 3D assets, it generalizes effectively to real 3D assets at inference time.
\vspace{-0.1cm}
\subsection{Preliminary}
\vspace{-0.1cm}
\label{sec:formulation}

The goal of native 3D texture generation is to predict colors directly in 3D space according to the given geometry and multi-view images. The geometric structure can be represented as
    $\G
    =
    \left\{
        \left(
            \mathbf{x}_i,
            \mathbf{n}_i,
            \mathbf{v}_i
        \right)
    \right\}_{i=1}^{N}$,
where $\mathbf{x}_i\in\R^3$ is the position of each point defined by the geometry, $\mathbf{n}_i\in\mathbb{S}^2$ is its normal vector, and $\mathbf{v}_i$ is its coordinate on a sparse voxel grid \citep{lai2025natex}. The corresponding texture is represented by normalized RGB colors
$
    \C
    =
    \left\{
        \mathbf{c}_i
    \right\}_{i=1}^{N},
$
where
$
    \mathbf{c}_i\in[-1,1]^3.
$
Given the target geometry $\G$ and a set of reference images $\I$, the model directly predicts the color at each 3D spatial location defined by the geometry, i.e.,
$
    p_{\theta}(\C\mid\G,\I).
    \label{eq:task}
$
Existing native texture models are typically trained on textured 3D assets that provide paired geometry and surface colors $(\G,\C)$, with multi-view conditioning images obtained by rendering the same assets.
\vspace{-0.1cm}
\subsection{From 2D Image to 3D Texture Supervision}
\vspace{-0.1cm}
\label{sec:data}

\begin{figure}[t]
    \centering
    \includegraphics[width=1.1\linewidth]{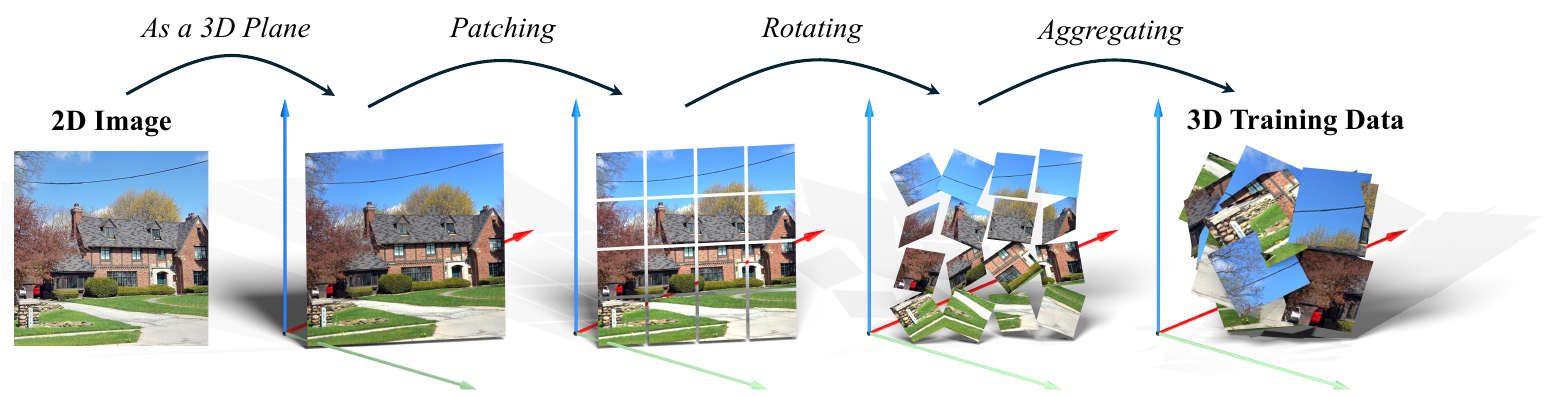}
    \caption{\textbf{Constructing 3D training data from 2D images.}
    We treat an image as a colored plane in 3D space, divide it into patches, and independently rotate and arrange these patches to construct a 3D sample with complex geometric structures.}
    \label{fig:data}
    \vspace{-1em}
\end{figure}

Large-scale 3D asset data are scarce and costly to acquire, limiting the availability and scale of training data for 3D texture generation. In contrast, vast amounts of high-quality 2D image data are readily available. In this work, we explore whether abundant 2D images can serve as an alternative source of training data for native 3D texture generation. To this end, we first develop a data preprocessing pipeline that converts each 2D image into an effective training sample that provides useful information required for native 3D texture model training, as illustrated in Figure~\ref{fig:data}.

\textbf{Image as a colored plane.} We first treat each 2D image as a colored plane in 3D space, converting each pixel into a point in 3D space. The coordinates of each pixel in the 2D image determine
the position $\mathbf{x}_i$ in 3D space, and its RGB value determines
the normalized color $\mathbf{c}_i$.
After coordinate normalization, all samples lie on the plane $z=0$
and share the unit normal $\mathbf{n}_i=(0,0,1)^\top$.
This mapping preserves the complete spatial and color information of the original 2D image while converting it into a 3D representation that can be directly processed by existing 3D texture models, forming the basis of the entire data construction process.

\textbf{Patch-wise geometric augmentation.} The geometry of a single plane is too simple to support effective learning of the complex geometric relationships found in real 3D surfaces. To introduce more complex geometry, we directly divide the image plane into non-overlapping patches and independently rotate each patch in 3D space. For a point $\mathbf{x}_i$ in patch $k$ with center $\boldsymbol{\mu}_k$, the transformation is 
\begin{equation} 
\widetilde{\mathbf{x}}_i = R_k \left( \mathbf{x}_i-\boldsymbol{\mu}_k \right) + \mathbf{t}_k, \qquad \widetilde{\mathbf{n}}_i = R_k\mathbf{n}_i, \qquad \widetilde{\mathbf{c}}_i = \mathbf{c}_i, \label{eq:transform} 
\end{equation} 
where $R_k$ is a randomly sampled 3D rotation matrix and $\mathbf{t}_k$ represents the location of the patch center in 3D space. The same rotation is applied to all points and normals within each patch, while their RGB values remain unchanged. 

\textbf{Aggregation.} Although independent patch transformations introduce more complex geometry, the resulting samples always contain isolated, floating patches. This differs from the coherent overall structure typically found in real 3D assets. To reduce this mismatch, we aggregate the rotated patches within a bounded 3D region, encouraging spatial overlap rather than allowing them to remain widely separated. The resulting sample exhibits more complex geometric structure and view-dependent occlusions while preserving the content within each patch.

\textbf{Training sample construction.}
We assign the transformed surface points to a 3D voxel grid of resolution $R$. If multiple points fall into the same voxel, we keep only one of them, without changing its position, normal, or RGB color. The remaining points define the geometry and texture of the training sample, which can be formulated as \begin{equation} 
\widetilde{\G} = \left\{ \left( \widetilde{\mathbf{x}}_i, \widetilde{\mathbf{n}}_i, \widetilde{\mathbf{v}}_i \right) \right\}_{i=1}^{N}, \qquad \widetilde{\C} = \left\{ \widetilde{\mathbf{c}}_i \right\}_{i=1}^{N}, \end{equation} 
where $\widetilde{\mathbf{v}}_i$ is the grid coordinate of the voxel with point $\widetilde{\mathbf{x}}_i$, and $N$ is the number of points.

We further render the colored point cloud from canonical orthographic
viewpoints to obtain the conditioning images $\widetilde{\I}$ and their
foreground masks. Each source 2D image thus yields a training tuple
$(\widetilde{\G},\widetilde{\C},\widetilde{\I})$ with known
geometry-color correspondences.
The pair $(\widetilde{\G},\widetilde{\C})$ is used for training Tex-Zero VAE, while the full tuple is used to train the Tex-Zero DiT to generate
surface colors conditioned on geometry and multi-view images. We provide a more comprehensive analysis of why the proposed data construction is effective for training texture generation models in Appendix~\ref{app:theoretical_analysis}.

\subsection{Tex-Zero VAE}
\label{sec:vae}

\textbf{Model Design.}
Given the geometry $\G$ and its corresponding colors $\C$, the Tex-Zero VAE reconstructs the colors conditioned on geometric positions following \citep{lai2025natex} as
\begin{equation}
    Z = E(\C,\G); \quad
    \widehat{\C}
    =
    D
    \left(
        Z,
        \G
    \right),
    \label{eq:vae}
\end{equation}
where $E$ and $D$ denote the VAE encoder and decoder, $Z$ denotes the encoded latent and $\widehat{\C}$ denotes the reconstructed colors.

The network architecture of Tex-Zero VAE can be built upon any standard image VAE by replacing dense 2D operations with sparse 3D operations on surface features. In our implementation, we adopt an overall architecture similar to FLUX VAE in image domain \citep{labs2024flux} and implement its encoder and decoder using sparse 3D operations.
    
\textbf{Training.}
The Tex-Zero VAE is first trained with a warm-up stage. During warm-up, we represent each image as a single colored plane and apply a random 3D rotation to the plane, without patch-wise operations. This stage allows the 3D VAE to start from a relatively simple task, facilitating stable optimization. Without this warm-up stage, the VAE fails to converge during training.

After warming up, we adopt the full data construction pipeline described in Section~\ref{sec:data}, where each image is divided into patches that are independently rotated, and arranged in 3D space. The VAE is thus trained on more complex surface orientations, discontinuities, and visibility patterns.

Tex-Zero VAE is optimized using a pointwise color reconstruction loss $\mathcal{L}_{\mathrm{color}}$, an image-space perceptual loss $\mathcal{L}_{\mathrm{perc}}$, and KL regularization $\mathcal{L}_{\mathrm{KL}}$:
\begin{equation}
    \mathcal{L}_{\mathrm{VAE}}
    =
    \mathcal{L}_{\mathrm{color}}
    +
    \lambda_{\mathrm{perc}}
    \mathcal{L}_{\mathrm{perc}}
    +
    \lambda_{\mathrm{KL}}
    \mathcal{L}_{\mathrm{KL}}.
    \label{eq:vae_loss}
\end{equation}
To compute the perceptual loss, we invert the transformations applied to the patches and assemble the reconstructed colors back into the original 2D image layout. Then, we apply LPIPS between the reconstructed and original images as the perceptual loss.

\textbf{Inference.}
During inference, real textured 3D assets are fed into the Tex-Zero VAE, where surface colors are encoded into latent features and subsequently decoded conditioned on geometry. Although the model is trained solely on image data, it transfers effectively to real 3D texture reconstruction. 

Moreover, Tex-Zero VAE can also encode 2D images if they are represented as a colored plane in 3D space.
The use of large-scale, high-quality image data enables the VAE to reconstruct 2D images with high fidelity while preserving fine-grained appearance details, which is challenging for existing 3D VAEs trained on 3D asset data. This capability provides a new perspective to encode image conditions for 3D texture generation, allowing 2D images and 3D textures to be processed with the same VAE and thereby bridging the representation gap between the two modalities.

\begin{figure}[t]
    \centering
    \makebox[\linewidth][l]{%
        \hspace*{-0.5cm}%
        \includegraphics[width=1.03\linewidth]{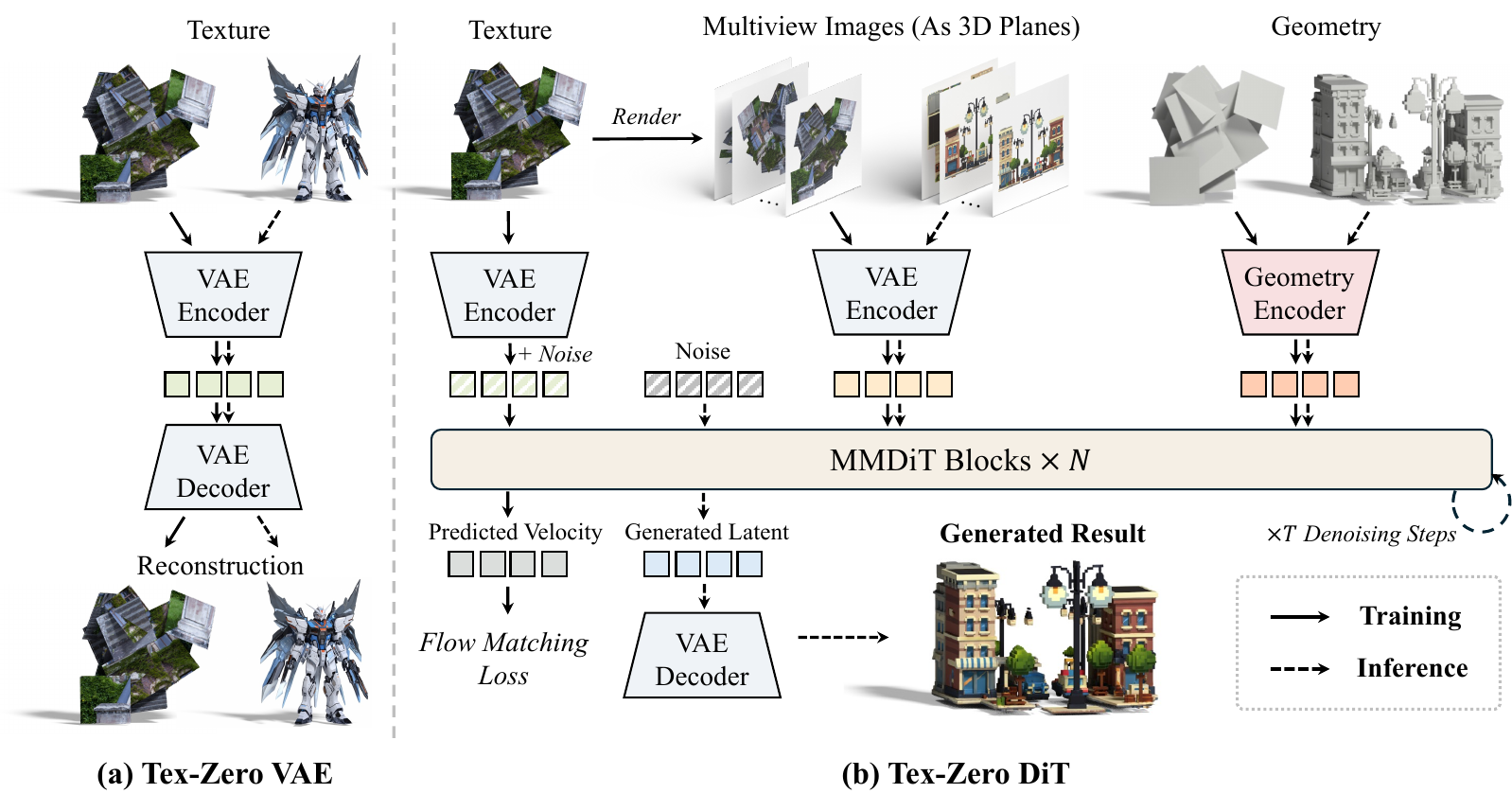}%
    }
    \caption{\textbf{Overview of Tex-Zero.}
    The Tex-Zero VAE encodes target textures and multi-view image conditions
    into a unified latent space. Conditioned on the image and geometry features,
    the Tex-Zero DiT generates high-fidelity 3D textures. Although the entire
    framework is trained exclusively on 2D images, it can directly generate
    textures for real 3D assets at inference time.}
    \label{fig:pipeline}
    \vspace{-1em}
\end{figure}

\subsection{Tex-Zero DiT}
\label{sec:dit}

\textbf{Model Design.}
The Tex-Zero DiT takes a noisy texture latent $Z_t$, the multi-view reference images $I$, and geometry as input. For image conditioning, we represent each of the six canonical views as a colored plane oriented according to its viewing direction, which are then encoded by the frozen Tex-Zero VAE. The resulting latent tokens are concatenated to form the conditioning sequence $Y$. For geometry conditioning, we encode the surface normals and concatenate the resulting features with $Z_t$ along the channel dimension following \citep{lai2025natex}. 

To represent spatial positions and distinguish different views, we design a joint rotary positional embeddings (RoPE). Specifically, we assign each token a four-dimensional positional index: 
\begin{equation} 
\mathbf{p}_j = \left( g_j,\, s_{j,x},\, s_{j,y},\, s_{j,z} \right), 
\label{eq:grouped_rope}
\end{equation} 
where $\mathbf{s}_j=(s_{j,x},s_{j,y},s_{j,z})$ denotes the three-dimensional latent-grid coordinates of each token. We set $g_j=0$ for target noisy 3D texture tokens and assign $g_j=\{1,\ldots,6\}$ to conditioning tokens from the front, right, back, left, top, and bottom views, respectively. We apply rotary embeddings independently along these four axes to separate channel groups of the attention queries and keys, allowing attention to incorporate both relative spatial positions and group identity.

The formulation of Tex-Zero DiT is compatible with any standard Multi-Modal DiT (MMDiT) backbones. We directly adopt the FLUX \citep{labs2024flux} MMDiT architecture. 

\textbf{Training.}
The Tex-Zero DiT is solely trained on the samples transformed from 2D images as described in Section~\ref{sec:data}, without using any real textured 3D assets. Rather than using all six rendered views, we randomly sample a subset of views at each training step, varying both the number and combination of conditioning views. This encourages the model to generate textures under different view-conditioning configurations rather than relying on a fixed set of views. Notably, some regions of each image patch within the constructed 3D sample may be occluded or absent from the conditioning images, requiring the model to predict their contents from the visible context. Since our data transformation process preserves the original content within each patch, its visible and occluded regions remain visually and semantically consistent, making this a natural task similar to outpainting. Such training can help the model to predict occluded regions when applied to real 3D assets.

We adopt the standard flow-matching objective \citep{lipman2023flow}. Given a target texture latent $Z$, Gaussian noise $\epsilon\sim\mathcal{N}(0,I)$, and $t\sim\mathcal{U}[0,1]$, we define $Z_t=(1-t)\epsilon+tZ$ and $u^{\star}=Z-\epsilon$.
The training loss can be represented as
\begin{equation}
    \mathcal{L}_{\mathrm{DiT}}
    =
    \mathbb{E}
    \left[
        \left\|
            f_{\theta}(Z_t;t;Y;G)-u^{\star}
        \right\|_F^2
    \right].
    \label{eq:flow_loss}
\end{equation}
We additionally apply random image-condition dropout to enable classifier-free guidance.

\textbf{Inference.}
At inference time, Tex-Zero takes a real 3D geometry and its multi-view images as input. We encode its multi-view images into condition features $Y$ and surface normals into geometry features $G$. Multi-step denoising is the adopted the obtain the clean texture latent. Finally, the frozen Tex-Zero VAE decoder maps the generated latent to RGB colors on the target surface.

\section{Experiments}
\label{sec:experiments}
\subsection{Experimental setup}
\label{sec:setup}

\begin{wrapfigure}{r}{0.48\textwidth}
    \centering
    \vspace{-1.0em}
    \includegraphics[width=\linewidth]{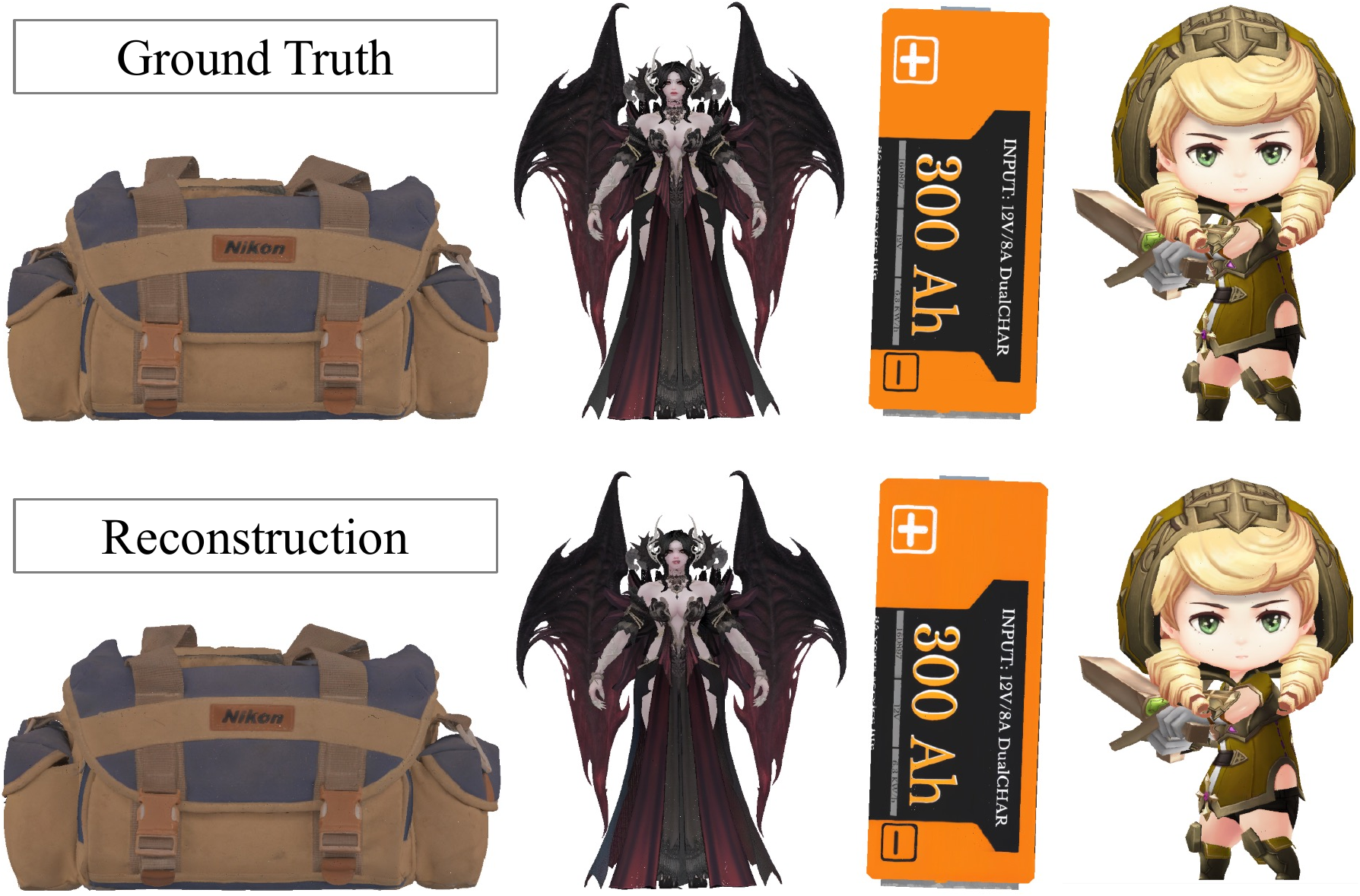}
    \caption{\textbf{Tex-Zero VAE Reconstruction Results.}
    Although trained without any 3D assets, the Tex-Zero VAE accurately encodes and reconstructs real 3D assets with fine-grained details.}
    \label{fig:vae_reconstruction}
\end{wrapfigure}

\textbf{Training settings.} We train both the Tex-Zero VAE and DiT on large-scale, publicly available image datasets, including SA-1B \citep{kirillov2023segment}, BLIP3o-60k \citep{chen2025blip3o} and ShareGPT-4o \citep{opengvlab2024sharegpt4o}, comprising approximately 11.1 million images in total. All images are resized to a resolution of $1536\times1536$ during training. Unless otherwise specified, the Tex-Zero DiT is trained based on the Tex-Zero VAE with a spatial downsampling factor of 16 and 16 latent channels. 
Appendix~\ref{app:details} specifies more detailed information for training.

\textbf{Comparisons and metrics.}
We primarily compare Tex-Zero with NaTex \citep{lai2025natex}, TRELLIS.2 \citep{xiang2025trellis2}, and our models trained on 3D data while keeping all other training settings unchanged. For the latter, we use approximately one million in-house textured 3D assets as the training set. Following prior work \citep{lai2025natex,chen2026lafite}, we evaluate texture quality on rendered views using PSNR, SSIM, and LPIPS. We also use the PSNR directly calculated on the point cloud (PSNR-PC) to evaluate the performance. More detailed information is provided in Appendix~\ref{app:setting}

\begin{table}[t]
    \centering
    \vspace{-1em}
    \caption{\textbf{VAE Reconstruction on 3D Textures and 2D Images.}
    $\mathrm{f}a\mathrm{c}b$ denotes a spatial downsampling factor of $a$
    and $b$ latent channels.}
    \label{tab:vae}
    \vspace{0.5em}
    \small
    \setlength{\tabcolsep}{3.5pt}
    \renewcommand{\arraystretch}{1.08}
    \resizebox{\linewidth}{!}{%
    \begin{tabular}{l c cccc ccc}
        \toprule
        \multirow[c]{2}{*}[-1ex]{Model}
        & \multirow[c]{2}{*}[-1ex]{Training Data}
        & \multicolumn{4}{c}{3D Asset Reconstruction}
        & \multicolumn{3}{c}{2D Image Reconstruction} \\
        \cmidrule(lr){3-6}
        \cmidrule(lr){7-9}
        &
        & \rule{0pt}{2.6ex}LPIPS$\downarrow$
        & PSNR-PC$\uparrow$
        & PSNR$\uparrow$
        & SSIM$\uparrow$
        & LPIPS$\downarrow$
        & PSNR$\uparrow$
        & SSIM$\uparrow$ \\
        \midrule

        \multicolumn{9}{l}{\textit{Existing methods}} \\

        FLUX VAE
        & Image
        & --
        & --
        & --
        & --
        & 0.0209
        & 39.63
        & 0.933 \\

        NaTex VAE
        & 3D Asset
        & 0.0374
        & 31.79
        & 40.94
        & 0.980
        & 0.1977
        & 34.36
        & 0.918 \\

        TRELLIS.2 VAE
        & 3D Asset
        & 0.0272
        & 33.94
        & 42.17
        & 0.988
        & 0.1487
        & 33.81
        & 0.915 \\

        \addlinespace[2pt]
        \multicolumn{9}{l}{\textit{Sparse VAE}} \\

        VAE-f8c16
        & 3D Asset
        & 0.0208
        & 36.55
        & 44.71
        & 0.989
        & 0.1430
        & 36.71
        & 0.944 \\

        VAE-f16c32
        & 3D Asset
        & 0.0343
        & 33.67
        & 41.84
        & 0.981
        & 0.2401
        & 32.43
        & 0.880 \\

        VAE-f16c16
        & 3D Asset
        & 0.0345
        & 30.90
        & 38.97
        & 0.974
        & 0.2993
        & 29.27
        & 0.833 \\

        \addlinespace[2pt]
        \rowcolor{gray!10}
        \multicolumn{9}{l}{\textit{\textbf{Tex-Zero Sparse VAE (ours)}}} \\

        \rowcolor{gray!10}
        VAE-f8c16
        & Image
        & 0.0154
        & 34.03
        & 42.41
        & 0.987
        & 0.0217
        & 39.82
        & 0.956 \\

        \rowcolor{gray!10}
        VAE-f16c32
        & Image
        & 0.0229
        & 32.59
        & 40.96
        & 0.980
        & 0.1093
        & 35.09
        & 0.931 \\

        \rowcolor{gray!10}
        VAE-f16c16
        & Image
        & 0.0316
        & 30.10
        & 38.54
        & 0.975
        & 0.1792
        & 32.33
        & 0.858 \\

        \bottomrule
    \end{tabular}%
    }
\end{table}

\subsection{Reconstruction}
\label{sec:reconstruction}

Table~\ref{tab:vae} compares Tex-Zero VAE with existing methods and baselines with the same architecture as Tex-Zero VAE but trained on textured 3D assets (i.e., \textit{Sparse VAE} in Table~\ref{tab:vae}). Despite never observing real 3D assets during training, the Tex-Zero VAE achieves competitive reconstruction performance on reconstructing real 3D assets. As shown in Figure~\ref{fig:vae_reconstruction}, it can effectively reconstruct fine-grained appearance details such as small text.
For 2D image reconstruction, the Tex-Zero VAE consistently outperforms its 3D-trained counterparts and other baselines. This demonstrates that its latents effectively preserve the fine-grained appearance information of the 2D images, enabling it to serve as an image encoder to encode multi-view image conditions for the Tex-Zero DiT.

\subsection{Generation}
\label{sec:generation}
\begin{figure}[t]
    \centering
    \includegraphics[width=0.94\linewidth]{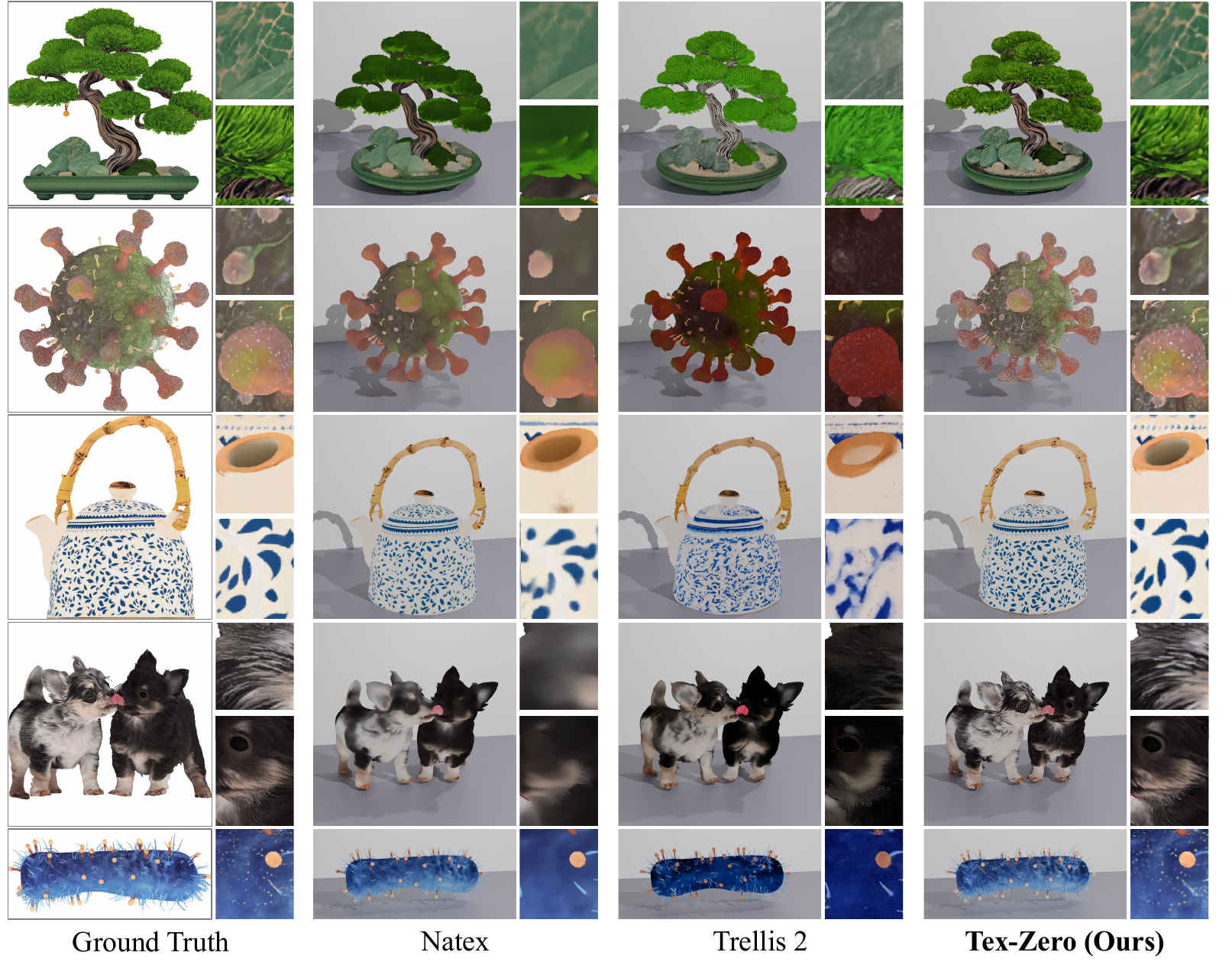}
    \vspace{-1em}
    \caption{\textbf{Visual Comparison between Tex-Zero and Baselines.} Tex-Zero can generate high-fidelity textures with fine-grained details.}
    \label{fig:main_results}
\end{figure}

\begin{table}[t]
    \centering
    \caption{\textbf{Quantitive Results for Texture Generation.}
    We compare Tex-Zero with two representative baselines, NaTex \citep{lai2025natex} and TRELLIS.2 \citep{xiang2025trellis2}.}
    \label{tab:generation}
    \small
    \setlength{\tabcolsep}{7pt}
    \renewcommand{\arraystretch}{1.08}
    \begin{tabular}{lrrrrrr}
        \toprule
        \multirow[c]{2}{*}[-0.5ex]{Method}
        & \multicolumn{3}{c}{Six-view}
        & \multicolumn{3}{c}{Front-view} \\
        \cmidrule(lr){2-4}
        \cmidrule(lr){5-7}
        & LPIPS$\downarrow$
        & PSNR$\uparrow$
        & SSIM$\uparrow$
        & LPIPS$\downarrow$
        & PSNR$\uparrow$
        & SSIM$\uparrow$ \\
        \midrule
        NaTex
        & 0.0754 & 27.74 & 0.949
        & 0.0669 & 27.49 & 0.947 \\
        TRELLIS.2
        & -- & -- & --
        & 0.1187 & 21.38 & 0.900 \\
        \rowcolor{gray!10}
        \textbf{Tex-Zero}
        & \textbf{0.0340} & \textbf{35.68} & \textbf{0.983}
        & \textbf{0.0294} & \textbf{35.31} & \textbf{0.985} \\
        \bottomrule
    \end{tabular}
\end{table}

We mainly compare Tex-Zero with two representative methods, NaTex and TRELLIS.2, in Table~\ref{tab:generation}. Tex-Zero consistently outperforms both baselines across all reported metrics under the six-view and front-view evaluation settings. As shown in Figure~\ref{fig:main_results}, Tex-Zero generates substantially finer texture details and more faithfully transfers the appearance information from the reference images to the 3D surface. These qualitative and quantitative results demonstrate the effectiveness of Tex-Zero for high-fidelity native 3D texture generation.

\subsection{2D \& 3D Joint Training }
\label{sec:mixed}

Image-derived data remains beneficial when textured 3D assets are available. Jointly training the VAE with 2D and 3D data effectively improves the reconstruction quality of the VAE, outperforming the model trained on 3D data alone (Table \ref{tab:mixed}). We further train the DiT using the mixed-data VAE and observe a similar improvement, demonstrating that when 3D training data are available, incorporating 2D data still provides additional improvements to both texture representation and generation.

\begin{table}[t]
    \centering
    \caption{\textbf{Joint Training with 2D and 3D Data.}
    Combining 2D images with textured 3D assets improves both VAE reconstruction and DiT generation.}

    \label{tab:mixed}
    \small
    \setlength{\tabcolsep}{7pt}
    \renewcommand{\arraystretch}{1.08}
    \begin{tabular}{l ccc ccc}
        \toprule
        \multirow{2}{*}{Training Data}
        & \multicolumn{3}{c}{VAE-f16c32}
        & \multicolumn{3}{c}{DiT} \\
        \cmidrule(lr){2-4}
        \cmidrule(lr){5-7}
        & LPIPS$\downarrow$
        & PSNR$\uparrow$
        & SSIM$\uparrow$
        & LPIPS$\downarrow$
        & PSNR$\uparrow$
        & SSIM$\uparrow$ \\
        \midrule
        3D
        & 0.0343
        & 41.84
        & 0.981
        & 0.0302
        & 36.23
        & 0.980 \\
        \rowcolor{gray!10}
        \textbf{2D + 3D}
        & \textbf{0.0140}
        & \textbf{42.85}
        & \textbf{0.988}
        & \textbf{0.0216}
        & \textbf{37.68}
        & \textbf{0.983} \\
        \bottomrule
    \end{tabular}
\end{table}

\vspace{1em}
\subsection{Ablation Study}

\label{sec:ablations}

\begin{wrapfigure}{r}{0.48\textwidth}
    \centering
    \vspace{-2em}
    \includegraphics[width=\linewidth]{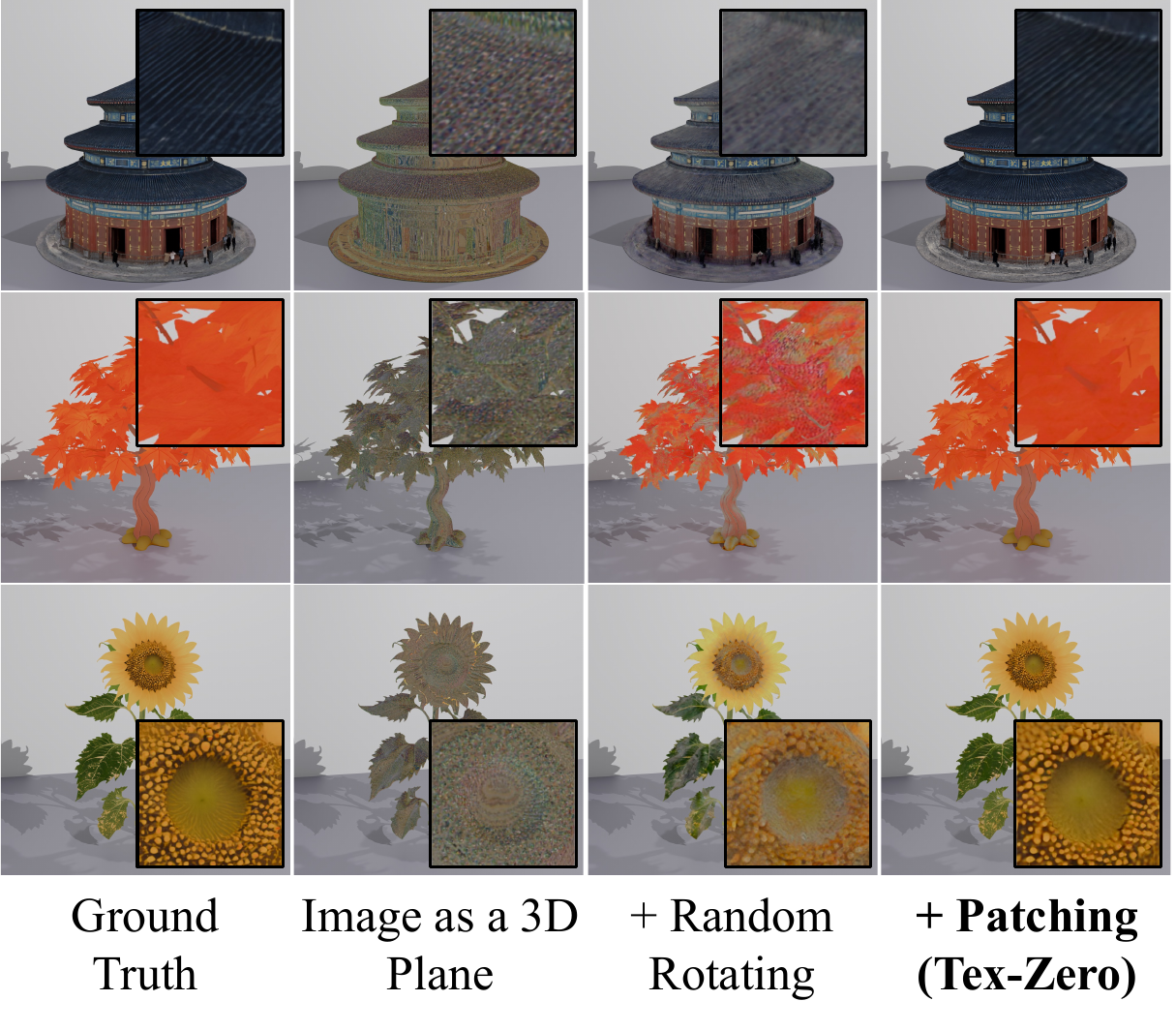}
    \vspace{-2.0em}
    \caption{\textbf{Ablation of the strategy for data preprocessing.}}
    \label{fig:vae_data_ablation}
    \vspace{-3em}
\end{wrapfigure}
\textbf{Ablation of Data Preprocessing.}
We train each VAE variant for 20K steps to evaluate the contribution of each preprocessing operation proposed in Section \ref{sec:data}. 
As shown in Table~\ref{tab:data_ablation} and Figure \ref{fig:vae_data_ablation}, 
training on a plane with a fixed position and orientation fails to transfer to real 3D assets. Since all training samples occupy a restricted planar neighborhood, many out-of-plane parameters in the sparse 3D operations cannot receive meaningful supervision. 
Randomly rotating the entire plane exposes the VAE to different surface orientations and enables meaningful 3D reconstruction, but obvious visible artifacts remain because the geometric structure of the training sample is too simple, resulting in a substantial gap from the distribution of geometry for real 3D assets. Applying independent rotations to image patches introduces more complex geometric structures, substantially improving reconstruction quality. Finally, aggregating the transformed patches into a compact structure reduces floaters and better approximates the spatial complexity of real surfaces, leading to further improvements. With 16 aggregated patches, the VAE achieves the best across all metrics (Table~\ref{tab:data_ablation}). More analysis is provided in Appendix \ref{app:why_vae}.

\begin{figure}[t]
    \centering
    \includegraphics[width=\linewidth]{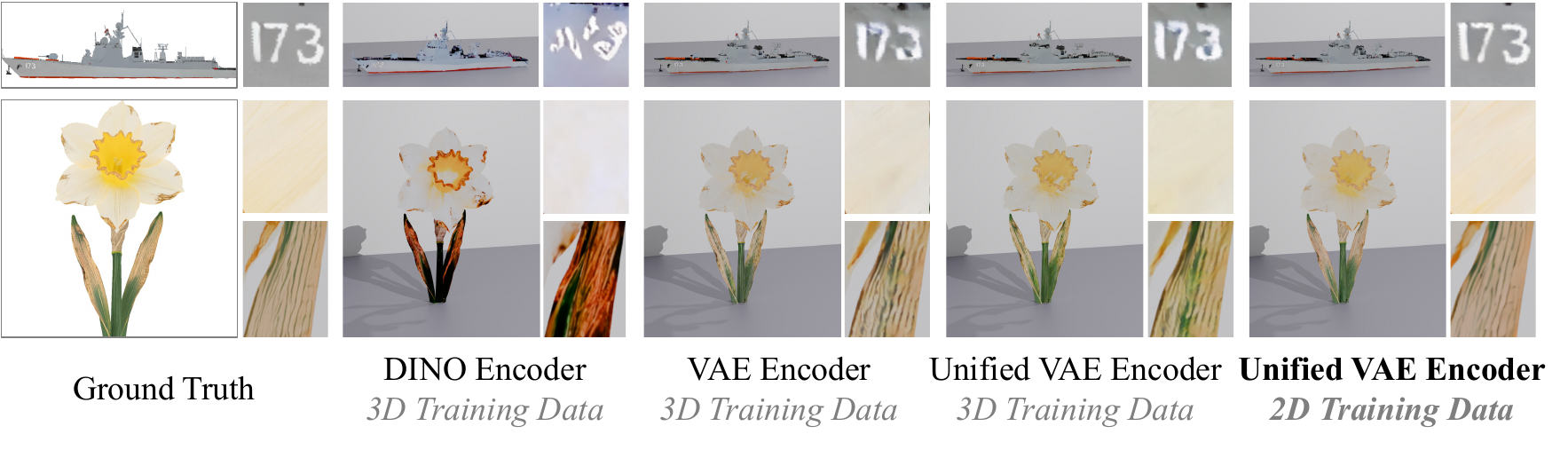}
    \caption{\textbf{Visualization of DiT generation results with different conditioning strategies.}
    Our image-trained Tex-Zero with the 2D-3D unified latent space best preserves fine-grained details from the reference images, achieving high-fidelity texture generation.}
    \label{fig:conditioning}
\end{figure}

\textbf{Ablation of the generation pipeline.}
We evaluate the effectiveness of the unified 2D-3D encoding for texture generation in Table~\ref{tab:conditioning} and Figure~\ref{fig:conditioning}. Specifically, we train the DiT under different conditioning injection strategies for 20k steps and evaluate their performance. Using the image-trained DINO or an independently trained VAE to encode the conditioning images leads to noticeable artifacts and loses fine-grained details from the reference images. 
Using the same 3D-trained VAE to encode the conditioning image improves generation by reducing this representation gap. However, the resulting textures remain blurry because a VAE trained only on 3D data has limited image reconstruction quality and discards details before they are passed to the DiT. The complete Tex-Zero pipeline trains both the shared VAE and DiT using 2D image data, achieving the best quantitative results and improving the fidelity of generated details.

\begin{table}[t]
    \centering
    \caption{\textbf{Ablation of data preprocessing.}
    We progressively introduce global rotation, patch-wise rotation, and spatial aggregation to explore their effect.}

    \label{tab:data_ablation}
    \small
    \setlength{\tabcolsep}{5pt}
    \begin{tabular}{ccc|cccc}
        \toprule
        Number of Patches & Random Rotation & Aggregation
        & LPIPS$\downarrow$
        & PSNR-PC$\uparrow$
        & PSNR$\uparrow$
        & SSIM$\uparrow$ \\
        \midrule
        1  & No  & No
        & 0.2160 & 11.34 & 18.16 & 0.777 \\
        1  & Yes & No
        & 0.1107 & 20.22 & 28.38 & 0.904 \\
        4  & Yes & No
        & 0.0591 & 26.83 & 35.05 & 0.963 \\
        4  & Yes & Yes
        & 0.0513 & 27.81 & 36.18 & 0.966 \\
        \rowcolor{gray!10}
        16 & Yes & Yes
        & \textbf{0.0355}
        & \textbf{28.97}
        & \textbf{37.16}
        & \textbf{0.973} \\
        \bottomrule
    \end{tabular}
\end{table}

\begin{table}[t]
    \centering

    \caption{\textbf{Effects of different conditioning strategies on DiT generation.} Unified 2D-3D VAE with 2D training data achieves the best results.}

    \label{tab:conditioning}
    \small
    \setlength{\tabcolsep}{7pt}
    \renewcommand{\arraystretch}{1.08}
    \begin{tabular}{lcccccc}
        \toprule
        Image Encoder
        & Training Data
        & Unified Latent
        & LPIPS$\downarrow$
        & PSNR$\uparrow$
        & SSIM$\uparrow$ \\
        \midrule
        DINO
        & Image
        & No
        & 0.1228
        & 25.67
        & 0.886 \\

        Separate VAE
        & Image
        & No
        & 0.0838
        & 29.65
        & 0.955 \\

        Shared VAE
        & 3D Asset
        & Yes
        & 0.0835
        & 30.42
        & 0.966 \\

        \rowcolor{gray!10}
        Tex-Zero VAE
        & Image
        & Yes
        & \textbf{0.0481}
        & \textbf{32.54}
        & \textbf{0.970} \\
        \bottomrule
    \end{tabular}
\end{table}

\section{Conclusion}
\label{sec:conclusion}
In this work, we propose Tex-Zero, a high-fidelity native 3D texture generation framework trained without any textured 3D assets. By converting 2D images into geometrically diverse samples in 3D space, Tex-Zero enables both its VAE and DiT to be trained entirely from large-scale image data. The Tex-Zero VAE reconstructs both real 3D textures and 2D images with high performance. Based on Tex-Zero VAE, Tex-Zero DiT is also trained solely from images, with a unified latent space of multi-view images and target textures. Experiments demonstrate that Tex-Zero generates high-quality textures with fine-grained details and outperforms several representative baselines, suggesting a new paradigm for constructing training data for 3D texture generation.

\label{sec:main_end}

\bibliography{iclr2027_conference}
\bibliographystyle{colm2024_conference}

\clearpage
\appendix

\section{More Explanations on Data Construction}
\label{app:theoretical_analysis}

We explain why the constructed samples provide valid supervision for both the 3D VAE and the conditional DiT. Our central observation is that the two roles of a textured 3D asset can be decoupled: its appearance provides the color signal to be modeled, whereas its geometry determines how this signal is organized in 3D space and observed across views. Neither module is trained to generate geometry itself. The VAE learns to represent colors over geometry-defined sparse neighborhoods, while the DiT learns how geometry organizes and partially reveals appearance through multi-view observations. This suggests that realistic object geometry is not intrinsically required for training. Instead, high-quality images can provide the appearance supervision, while constructed geometries provide the directional, spatial, and visibility interactions required by the models. The following analysis shows how our patch-wise transformations and spatial aggregation instantiate these task-relevant interactions without requiring semantically meaningful 3D assets.

\subsection{Converting Image Patches into 3D Training Samples}
\label{app:image_patches}

We first represent each image pixel as a colored point on the plane
\(z=0\):
\begin{equation}
    \mathbf{x}_i
    =
    (u_i,v_i,0)^\top,
    \qquad
    \mathbf{n}_i
    =
    (0,0,1)^\top,
    \qquad
    \mathbf{c}_i
    \in[-1,1]^3,
    \label{eq:app_planar_points}
\end{equation}
where \((u_i,v_i)\) is the pixel coordinate, \(\mathbf{n}_i\) is the
surface normal, and \(\mathbf{c}_i\) is the corresponding RGB value.

For every point \(i\) in patch \(k\), we apply the same rotation and
translation:
\begin{equation}
    \widetilde{\mathbf{x}}_i
    =
    R_k
    \left(
        \mathbf{x}_i-\boldsymbol{\mu}_k
    \right)
    +
    \mathbf{t}_k,
    \qquad
    \widetilde{\mathbf{n}}_i
    =
    R_k\mathbf{n}_i,
    \qquad
    \widetilde{\mathbf{c}}_i
    =
    \mathbf{c}_i,
    \label{eq:app_constructed_patch}
\end{equation}
where \(R_k\in\mathrm{SO}(3)\), \(\boldsymbol{\mu}_k\) is the patch
center, and \(\mathbf{t}_k\) determines its position in 3D space.

Since a rotation preserves Euclidean distances, any two points \(i\) and
\(j\) in the same patch satisfy
\begin{equation}
\begin{aligned}
    \left\|
        \widetilde{\mathbf{x}}_i
        -
        \widetilde{\mathbf{x}}_j
    \right\|_2
    &=
    \left\|
        R_k
        \left(
            \mathbf{x}_i-\mathbf{x}_j
        \right)
    \right\|_2 \\
    &=
    \left\|
        \mathbf{x}_i-\mathbf{x}_j
    \right\|_2.
\end{aligned}
\label{eq:app_patch_distance}
\end{equation}
Meanwhile, their colors remain unchanged:
\begin{equation}
    \widetilde{\mathbf{c}}_i=\mathbf{c}_i.
    \label{eq:app_color_preservation}
\end{equation}

Therefore, the transformation changes only the position and orientation
of each patch in 3D space, while preserving its internal spatial
structure and appearance content. In particular, the original colors,
edges, and fine-grained texture patterns remain intact. After
voxelization, the transformed points form a sparse 3D support with a
well-defined color at every occupied location, and can therefore be used
directly as training data for the 3D VAE.

\subsection{Why the Data Can Train the 3D VAE}
\label{app:why_vae}

The 3D VAE encodes and reconstructs surface colors on a given geometric support:
\begin{equation}
    Z=E(\C,\G),
    \qquad
    \widehat{\C}=D(Z,\G).
    \label{eq:app_vae_mapping}
\end{equation}
It does not predict the geometry \(\G\). Instead, \(\G\) specifies the occupied voxels and their spatial neighborhoods, while the reconstruction target is the color field \(\C\). A valid training sample therefore requires a three-dimensional support with a well-defined color at every occupied location, which is exactly provided by the transformed image patches in Equation~\eqref{eq:app_constructed_patch}.

The geometric transformations prevent these sparse supports from degenerating into a restricted subset of three-dimensional configurations. Consider a sparse convolution
\begin{equation}
    y_{\mathbf v}
    =
    \sum_{\boldsymbol{\Delta}\in\mathcal K}
    \mathbf 1
    \left\{
        \mathbf v+\boldsymbol{\Delta}\in\mathcal V
    \right\}
    W_{\boldsymbol{\Delta}}
    x_{\mathbf v+\boldsymbol{\Delta}},
    \label{eq:app_sparse_conv}
\end{equation}
where \(\mathcal V\) is the occupied voxel set and
\(\mathcal K\subset\mathbb Z^3\) is the convolutional kernel support.

\paragraph{Directional Degeneracy of Fixed-Plane Training.}
Suppose that every training sample lies on the fixed plane \(v_z=0\).
For an occupied output voxel \(\mathbf v\) and any offset satisfying
\(\Delta_z\neq0\),
\begin{equation}
    \mathbf 1
    \left\{
        \mathbf v+\boldsymbol{\Delta}\in\mathcal V
    \right\}
    =0.
    \label{eq:app_inactive_offset}
\end{equation}
Consequently, the output of Equation~\eqref{eq:app_sparse_conv} is
independent of \(W_{\boldsymbol{\Delta}}\), and
\begin{equation}
    \frac{\partial\mathcal L}
    {\partial W_{\boldsymbol{\Delta}}}
    =0
    \label{eq:app_zero_gradient}
\end{equation}
for any submanifold sparse-convolution layer whose active support remains
on the plane. A fixed plane therefore cannot provide supervision for all
three-dimensional kernel directions. Randomly rotating the plane changes
which voxel offsets are activated and provides directional supervision for
the 3D operators.

\paragraph{Coplanarity of Globally Rotated Samples.}
A globally rotated plane nevertheless remains coplanar. Let
\begin{equation}
    \Sigma_x
    =
    \frac{1}{N}
    \sum_{i=1}^{N}
    (\mathbf x_i-\bar{\mathbf x})
    (\mathbf x_i-\bar{\mathbf x})^\top
    \label{eq:app_coordinate_covariance}
\end{equation}
be the coordinate covariance matrix of a sample. Every globally rotated
plane satisfies
\begin{equation}
    \lambda_{\min}(\Sigma_x)=0,
    \label{eq:app_planar_rank}
\end{equation}
whereas a general non-coplanar support need not satisfy this identity.
Independently rotating the patches removes this restriction and allows a
single training sample to contain multiple local surface orientations.

\paragraph{Spatial Aggregation of Independently Rotated Patches.}
Independent rotations alone do not guarantee that differently oriented
patches interact spatially. If the patches remain far apart, a local 3D
operator processes them as separate planar components. Spatial aggregation
places the transformed patches within a compact region, allowing them to
intersect or approach one another.

Let \(P_k\subset\mathbb R^3\) denote the support of patch \(k\), and let
\(r_L\) be the effective spatial receptive radius of the 3D VAE. Whenever
\begin{equation}
    \operatorname{dist}(P_i,P_j)\le r_L,
    \label{eq:app_patch_interaction}
\end{equation}
some receptive fields contain occupied voxels from both patches. Their
features therefore depend jointly on surfaces with different orientations,
rather than being computed independently on isolated planes. A direct
intersection is the special case
\(\operatorname{dist}(P_i,P_j)=0\).

The three operations consequently address distinct geometric degeneracies:
\begin{equation}
\begin{aligned}
    \text{global rotation}
    &\Rightarrow
    \text{supervision along different 3D directions},\\
    \text{patch-wise rotation}
    &\Rightarrow
    \text{multiple surface orientations within one sample},\\
    \text{spatial aggregation}
    &\Rightarrow
    \text{interacting multi-patch neighborhoods}.
\end{aligned}
\label{eq:app_vae_roles}
\end{equation}

These results explain why
the constructed samples provide effective supervision for the 3D VAE.
Rigid transformations preserve the intrinsic texture content of each image
patch, while random rotations and spatial aggregation construct
directionally diverse, non-coplanar, and spatially interacting sparse
supports. The VAE can therefore learn to encode and reconstruct colors on
the types of three-dimensional neighborhoods encountered at inference,
without requiring the training samples to form semantically meaningful
object shapes.

\subsection{Why the Data Can Train the DiT}
\label{app:why_dit}

The DiT learns a geometry-conditioned mapping from multi-view observations with known viewpoints and a given geometry to a complete surface texture:
\begin{equation}
    F_{\theta}:
    \left(
        \G,\{\I_v\}_{v\in V}
    \right)
    \longmapsto
    \C,
    \label{eq:app_dit_mapping}
\end{equation}
where \(\G\) is the target geometry, \(\I_v\) is the conditioning image observed from viewpoint \(v\), and \(\C\) is the complete texture defined on \(\G\). The learning problem is therefore to integrate appearance information across views under the geometric constraints imposed by \(\G\), and to complete surface regions that are not observed in the conditioning views.

For a viewpoint \(v\), let \(\mathcal A_v(\G)\) denote the geometry-dependent observation operator induced by camera projection, depth ordering, and visibility. We write
\begin{equation}
    \I_v
    =
    \mathcal A_v(\G)\C
    +
    \boldsymbol{\eta}_v,
    \label{eq:app_observation_operator}
\end{equation}
where \(\boldsymbol{\eta}_v\) accounts for image components that are not fully explained by the target geometry, such as local misalignment, background content, or view-dependent artifacts. For fixed \(\G\), the operator \(\mathcal A_v(\G)\) determines how surface regions are projected, which regions are visible, and how visibility is resolved when multiple surfaces overlap in projection.

Our constructed data follow the same geometry-conditioned structure. Patch-wise transformations assign different spatial positions and orientations to the image patches. Their subsequent spatial arrangement allows multiple patches to overlap under projection and to occlude one another. Let \(\Pi_{k,v}\) denote the projected region of patch \(k\) under viewpoint \(v\), and let \(d_{k,v}(p)\) be its depth at pixel \(p\). When \(p\) is covered by multiple projected patches, the visible patch is determined by
\begin{equation}
    k_v^\star(p)
    =
    \operatorname*{arg\,min}_{k:\,p\in\Pi_{k,v}}
    d_{k,v}(p).
    \label{eq:app_visible_patch}
\end{equation}
Accordingly, the visible region of patch \(k\) is
\begin{equation}
    \Omega_{k,v}
    =
    \left\{
        p\in\Pi_{k,v}:
        k=k_v^\star(p)
    \right\}.
    \label{eq:app_visible_region}
\end{equation}
The partition
\(\{\Omega_{k,v}\}_{k=1}^{K}\) is thus jointly determined by the patch positions, orientations, camera viewpoint, and depth ordering.

For two overlapping patches \(i\) and \(j\), their visibility switches along the depth-equality set
\begin{equation}
    B_{ij,v}
    =
    \left\{
        p:
        d_{i,v}(p)=d_{j,v}(p)
    \right\}.
    \label{eq:app_visibility_boundary}
\end{equation}
Whenever
\begin{equation}
    \nabla
    \left(
        d_{i,v}-d_{j,v}
    \right)
    \neq 0,
\end{equation}
the implicit function theorem implies that \(B_{ij,v}\) is locally a regular image-space curve. Crossing this curve changes the frontmost patch from \(i\) to \(j\), or vice versa. Since different patches generally contain different image content, the resulting observation contains appearance regions and boundaries whose spatial organization is determined by the underlying geometry.

The same projection and visibility rules produce silhouettes, self-occlusion boundaries, and transitions between visible surface regions on real 3D assets. The constructed data therefore convert three-dimensional spatial relations into observable geometry-dependent structures in the conditioning images. As the patch arrangement and viewpoint vary, the projected regions, depth ordering, and occlusion patterns vary accordingly. This provides the DiT with diverse examples of how a given geometry organizes multi-view appearance and how information from different views should be integrated on the target surface.

The construction also provides supervision for completing unobserved regions. Let \(A_k\) denote the complete appearance field of source-image patch \(k\), and let
\begin{equation}
    M_{k,V}
    =
    M_{k,V}(\widetilde{\G})
    \label{eq:app_visibility_mask}
\end{equation}
denote its visibility mask under the constructed geometry
\(\widetilde{\G}\) and the selected viewpoints \(V\). Its visible and hidden parts are
\begin{equation}
    A_{k,\mathrm{vis}}
    =
    M_{k,V}\odot A_k,
    \qquad
    A_{k,\mathrm{hid}}
    =
    \left(1-M_{k,V}\right)\odot A_k.
    \label{eq:app_visible_hidden}
\end{equation}

The rigid transformation of a patch changes only its position and orientation in 3D space; it preserves the complete appearance and spatial organization of \(A_k\). Consequently, \(A_{k,\mathrm{vis}}\) and \(A_{k,\mathrm{hid}}\) are complementary subsets of the same coherent appearance field rather than independently sampled content. In particular, the construction preserves their joint image statistics,
\begin{equation}
    P_{\mathrm{img}}
    \left(
        A_{k,\mathrm{vis}},
        A_{k,\mathrm{hid}}
        \mid
        M_{k,V}
    \right),
    \label{eq:app_joint_appearance}
\end{equation}
and presents the model with the conditional completion problem
\begin{equation}
    P_{\mathrm{img}}
    \left(
        A_{k,\mathrm{hid}}
        \mid
        A_{k,\mathrm{vis}},
        M_{k,V}
    \right).
    \label{eq:app_completion_distribution}
\end{equation}

This has the same statistical form as image outpainting, except that the missing regions are induced by three-dimensional projection and occlusion rather than by an independently sampled two-dimensional mask. By varying the patch arrangement and the selected conditioning views, the construction generates diverse visibility masks,
\begin{equation}
    M_{k,V}
    \sim
    P_M
    \left(
        M
        \mid
        \widetilde{\G},V
    \right).
    \label{eq:app_mask_distribution}
\end{equation}
For each such mask, the conditioning observations contain only part of the appearance, whereas the training target contains the complete texture. The training objective therefore encourages the DiT to use the semantic and textural context in visible regions to model plausible appearance in unobserved regions. When the hidden content is not uniquely determined by the observations, the appropriate target is its conditional distribution rather than a deterministic recovery rule.

The constructed samples consequently provide two coupled forms of supervision. First, they require multi-view integration: the model must combine observations whose organization changes with projection, depth ordering, and visibility under the supplied geometry. Second, they require geometry-conditioned completion: the model must predict surface appearance that is absent from the selected views while remaining consistent with the visible context.

These are the same two components of the inference problem on real assets:
\begin{equation}
    \left(
    \begin{gathered}
        \text{given geometry,}\\
        \text{geometry-organized multi-view observations}
    \end{gathered}
    \right)
    \longmapsto
    \text{complete surface texture}.
    \label{eq:app_common_task}
\end{equation}
Accordingly, the validity of the supervision does not require the constructed geometry to resemble a semantically meaningful object. Patch-wise transformations and spatial aggregation instead provide diverse surface orientations, projections, depth orderings, occlusion patterns, and geometry-dependent appearance organizations. At the same time, preserving the complete content within each patch retains the statistical relationship between visible and hidden appearance. The resulting data therefore instantiate the multi-view integration and conditional completion problems required by the DiT, allowing the learned mapping to transfer to real geometries governed by the same projection and visibility mechanisms.

\subsection{Conclusion.}
Image-derived data provide effective supervision for the two components in complementary ways. For the 3D VAE, rigid patch transformations preserve the intrinsic texture content while producing diverse directional, non-coplanar, and spatially compact sparse supports for learning geometry-conditioned color representations. For the DiT, patch transformations and multi-view rendering generate varied projection, depth-ordering, and visibility patterns, causing the observed appearance to be organized by the constructed geometry. Moreover, because only geometry-dependent subsets of each patch are visible in the conditioning views while its complete content remains in the target, the training data naturally induce a geometry-conditioned completion task analogous to image outpainting. Therefore, the usefulness of the constructed data does not rely on reproducing semantically realistic object shapes. Instead, it preserves the original semantic and textural relationships between the visible and occluded regions within each image patch, while exposing the models to the 3D support structures and observation mechanisms required for real 3D texturing.

\section{Implementation Details}
\label{app:details}

This appendix provides additional details of the data representation,
VAE architecture, DiT conditioning, and optimization.
Unless otherwise specified, the network settings below refer to the
f16c16 VAE and its corresponding DiT.
We use $S$ to denote image resolution, $R$ to denote voxel-grid
resolution, and $L$ to denote the number of latent tokens.


\subsection{Data Representation and Preprocessing}
\label{app:data}

\paragraph{Image preprocessing.}
Images are converted to RGB, center-cropped to a square, and resized
using Lanczos resampling.
RGB values are normalized to $[-1,1]$.
Each pixel is associated with a point on the plane $z=0$, with its
horizontal and vertical coordinates normalized to the reference
range $[-1,1]$.
Pixel centers are used to determine point positions, and all points
initially have normal $(0,0,1)^\top$.
Thus, the initial point cloud preserves the spatial arrangement
and colors of the preprocessed image.

Image resolution and voxel-grid resolution serve different purposes.
The former determines the number of source pixels, whereas the latter
determines the spatial discretization used by the sparse network.
The VAE configuration uses $S=1536$ and $R=1536$.
The DiT configuration constructs samples from images with $S=1536$
and encodes them at $R=1536$; its conditioning images are rendered
at $1536\times1536$.


\paragraph{Patch-wise rotation and aggregation.}
The image plane is divided into a regular grid of non-overlapping
patches.
For each patch, we independently sample rotations about the three
coordinate axes, with each angle drawn uniformly from
$[0,360^\circ)$.
The rotation is applied around the patch center, and a translation
determines its position in the constructed sample.
The DiT configuration uses a $4\times4$ grid, giving 16 patches
per image.

In the aggregation-enabled setting, patches are placed in a randomly
shuffled order.
Their positions are sampled so that the axis-aligned bounding box
of each new patch overlaps at least one previously placed box.
We allow up to 128 placement trials.
If these trials are unsuccessful, a guided placement is sampled
relative to an existing box.

For DiT training, the placement region is controlled by a half-size
$b$, sampled uniformly from $[0.75,1.0]$ for each example and shared
by all its patches.
Translations are constrained to keep patch bounding boxes inside
$[-b,b]^3$ whenever feasible.
The fallback prioritizes spatial grouping when this constraint
conflicts with the overlap requirement.
Bounding-box overlap encourages compact arrangements, but does not
require the underlying surfaces to be physically connected or
free of intersections.

\paragraph{Voxelization.}
Transformed points are assigned to a voxel grid, and only the first
point assigned to each occupied voxel is retained.
Its position, normal, color, and patch identity are preserved;
colors from different points are not averaged.
The resulting number of points therefore depends on the image
content resolution and the sampled geometric arrangement.

For image-derived samples, all retained points are used rather than
subsampling a fixed-size point cloud.
Examples with fewer than 100,000 retained points are rejected.
The retained patch identities and transformations are also stored
for the image-space reconstruction loss described below.

\subsection{Tex-Zero VAE}
\label{app:vae}

\paragraph{Network architecture.}
The Tex-Zero VAE follows the encoder--decoder structure of the
FLUX VAE, implemented with sparse 3D operations.
The input and output features have three channels corresponding
to RGB values.
Geometry determines the occupied voxel locations; the VAE
reconstructs their colors rather than predicting geometry.
Surface normals are not concatenated with RGB features during
color reconstruction.

The encoder contains five resolution levels with channel widths
$[128,256,512,512,512]$.
Each level contains two residual blocks.
Four factor-two downsampling operations produce a spatial
compression factor of 16.
The decoder mirrors these resolution levels and uses three
residual blocks per level.

Each residual block uses sparse $3\times3\times3$ convolutions,
32-group normalization, and SiLU activations.
Both the encoder and decoder include a bottleneck consisting of
two residual blocks with a full self-attention block between them.
The bottleneck attention uses two heads.
Downsampling and upsampling are implemented with sparse
pixel-unshuffle and pixel-shuffle operations, respectively,
together with grouped channel projections and sparse convolutions.

The encoder predicts the mean and log variance of a diagonal
Gaussian posterior.
For f16c16, each latent token has 16 channels.
The log variance is clamped to $[-30,20]$ for numerical stability,
and latent samples are obtained using the reparameterization trick. The VAE is trained from scratch.

\begin{table}[t]
    \centering
    \caption{\textbf{Architecture of the f16c16 Tex-Zero VAE.}
    Channel widths are listed from the finest to the coarsest
    resolution level.}
    \vspace{1em}
    \label{tab:app_vae_architecture}
    \small
    \setlength{\tabcolsep}{7pt}
    \begin{tabular}{ll}
        \toprule
        Setting & Value \\
        \midrule
        Input / output channels & 3 / 3 \\
        Resolution levels & 5 \\
        Channel widths & $[128,256,512,512,512]$ \\
        Encoder residual blocks per level & 2 \\
        Decoder residual blocks per level & 3 \\
        Spatial downsampling factor & 16 \\
        Latent channels & 16 \\
        Bottleneck attention heads & 2 \\
        Normalization & GroupNorm, 32 groups \\
        Normalization epsilon & $10^{-6}$ \\
        Activation & SiLU \\
        \bottomrule
    \end{tabular}
\end{table}

\paragraph{Reconstruction objective.}
The color reconstruction term in Equation~\ref{eq:vae_loss}
is the mean squared error over all retained points and RGB channels:
\begin{equation}
    \mathcal{L}_{\mathrm{color}}
    =
    \frac{1}{3N}
    \sum_{i=1}^{N}
    \left\|
        \widehat{\mathbf{c}}_i-\mathbf{c}_i
    \right\|_2^2.
    \label{eq:app_color_loss}
\end{equation}
KL regularization is computed against a standard Gaussian prior
and averaged over latent tokens and channels.
We use $\lambda_{\mathrm{perc}}=0.1$ and
$\lambda_{\mathrm{KL}}=10^{-6}$.
The perceptual network is a frozen VGG-based LPIPS model
\citep{zhang2018lpips}.

\paragraph{Image-space perceptual loss.}
For image-derived samples, the stored patch transformations map
the retained points back to their original image-plane coordinates.
We render the predicted and reference colors at these same
coordinates to obtain two reassembled images.
Both images are rendered at $1024\times1024$ with a white background.
To reduce gaps introduced by voxelization, we apply two iterations
of hole filling with a $3\times3$ neighborhood before evaluating
LPIPS.

The reference image used by this loss is rendered from the retained
ground-truth colors using the same procedure as the prediction.
This ensures that both images have the same spatial sampling.
The perceptual loss therefore compares appearance reconstruction
rather than differences caused by voxelization.
It is evaluated in the original image layout, not across six views
of the transformed sample.

For 3D-data training settings, the loss instead compares predicted
and reference colors rendered from six canonical orthographic
directions and sums the valid per-view LPIPS terms.

\paragraph{Latent scaling.}
The VAE interface applies an affine transformation to posterior
latents before decoding:
\begin{equation}
    z_{\mathrm{scaled}}
    =
    a\left(z_{\mathrm{raw}}-b\right),
    \qquad
    a=0.3611,\quad b=0.1159.
    \label{eq:app_latent_scaling}
\end{equation}
The decoder applies the inverse transformation before its first
network layer.
Our online DiT training operates on raw posterior latents.
Accordingly, generated latents are transformed using
Equation~\ref{eq:app_latent_scaling} before being passed to the
VAE decoder interface.

\subsection{Tex-Zero DiT}
\label{app:dit}

\paragraph{Network architecture.}
We use a FLUX-style Transformer with 12 dual-stream blocks followed
by 24 single-stream blocks.
The hidden width is 1024, the number of attention heads is 8,
and the MLP expansion ratio is 4.
Each target texture token contains 16 channels.
The corresponding 16-channel normal feature is concatenated
along the channel dimension, resulting in a 32-channel input.
The output contains 16 channels and predicts the texture-latent
velocity.

Conditioning-image latents have 16 channels and are linearly
projected to width 1024.
A learned view embedding is added to each projected token.
The target and conditioning streams exchange information through
joint attention in the dual-stream blocks and are subsequently
concatenated for processing by the single-stream blocks.
Timestep embeddings modulate the Transformer blocks and the
output layer.
Only target tokens are retained for velocity prediction.

\begin{table}[t]
    \centering
    \caption{\textbf{Architecture and conditioning settings of
    the Tex-Zero DiT.}}
    \vspace{1em}
    \label{tab:app_dit_architecture}
    \small
    \setlength{\tabcolsep}{7pt}
    \begin{tabular}{ll}
        \toprule
        Setting & Value \\
        \midrule
        Texture / normal latent channels & 16 / 16 \\
        Concatenated input channels & 32 \\
        Output channels & 16 \\
        Image-condition latent channels & 16 \\
        Hidden width & 1024 \\
        Attention heads & 8 \\
        MLP expansion ratio & 4 \\
        Dual-stream / single-stream blocks & 12 / 24 \\
        Rotary axes & $(g,x,y,z)$ \\
        Rotary dimensions per head & $[16,40,40,32]$ \\
        Rotary frequency base & 10,000 \\
        Image-condition dropout probability & 0.1 \\
        \bottomrule
    \end{tabular}
\end{table}

\paragraph{Online target and condition encoding.}
The Tex-Zero VAE is frozen throughout DiT training.
For each sample, it separately encodes surface colors and normals
at the same voxel locations.
The two resulting latent sequences have matching spatial
coordinates.

Conditioning images are generated online by orthographically
rendering the colored point cloud from the front, right, back,
left, top, and bottom directions.
The renderer resolves visibility with a depth buffer and returns
both RGB images and foreground masks.
No additional illumination model or hole filling is applied
when constructing these conditions.

Foreground pixels with mask values greater than 0.5 are embedded
as points on the corresponding oriented image plane.
Their colors are normalized to $[-1,1]$ and encoded using the
same frozen VAE.
Only the resulting color latents are used as image-conditioning
tokens.
Background pixels are excluded, so the conditioning-sequence
length varies across samples.

The target and conditioning planes are both encoded at voxel-grid
resolution 1536.
Target appearance and normal latents are sampled from their
respective posteriors during training, whereas image conditions
use posterior means.
At inference time, normal conditions also use posterior means.


\paragraph{Positional encoding.}
We use the four-dimensional token indices defined in
Equation~\ref{eq:grouped_rope}.
The spatial coordinates of target tokens come from the target
surface's sparse latent grid.
Those of conditioning tokens come from the latent grids of the
oriented image planes.
These image-plane coordinates encode pixel layout and viewing
direction, rather than per-pixel depth on the target surface.

The group index is zero for target tokens and takes values
one through six for the six canonical conditioning views.
Within each attention head, the 128 query and key channels
are divided into groups of $[16,40,40,32]$ channels for the
group, $x$, $y$, and $z$ axes, respectively.
Rotary embeddings are applied independently to these channel
groups with frequency base 10,000.

The positional-encoding reference resolution is 1536.
Since target and condition encoding use the same resolution,
no additional spatial-coordinate rescaling is required in
this configuration.

\paragraph{Flow-matching objective.}
We sample a timestep uniformly from $[0,1]$ and construct the
linear noise-to-data path described in the main text.
The DiT predicts the velocity $Z-\epsilon$ from the noisy latent,
image conditions, and normal features.
The squared prediction error is averaged over all target
tokens and latent channels, with no additional timestep-dependent
loss weighting.

With probability 0.1, we drop the image appearance condition
while retaining the geometry condition.
Specifically, the projected image-condition features are zeroed,
while their token positions and group indices are retained.
The model is trained from scratch.

\subsection{Optimization and Sampling}
\label{app:optimization}

\paragraph{Optimization.}
Both the VAE and DiT are optimized using AdamW with base learning
rate $10^{-4}$, betas $(0.9,0.99)$, epsilon $10^{-6}$, and
weight decay $10^{-2}$.
Biases and normalization parameters are excluded from weight decay.
Both configurations use FP16 mixed-precision training.

The learning-rate schedule uses linear warm-up followed by cosine
decay.
Its initial, maximum, and minimum multipliers relative to the
base learning rate are $10^{-6}$, $1$, and $10^{-3}$, respectively.
The configured learning-rate warm-up lasts one optimizer update
for the VAE and 100 updates for the DiT.
This learning-rate warm-up is separate from the single-plane
VAE training stage described in Section~\ref{sec:vae}.

Each training process loads one example per micro-batch.
Because the number of occupied voxels varies across samples,
the data loader maintains a buffer organized by point-cloud size.
The VAE loader uses eight workers with prefetch factor four;
the DiT loader uses four workers with prefetch factor two.

\begin{table}[t]
    \centering
    \caption{\textbf{Optimization settings.}
    The micro-batch size is specified per training process,
    before gradient accumulation.}
    \vspace{1em}
    \label{tab:app_optimization}
    \small
    \setlength{\tabcolsep}{7pt}
    \begin{tabular}{lcc}
        \toprule
        Setting & VAE & DiT \\
        \midrule
        Optimizer & AdamW & AdamW \\
        Base learning rate & $10^{-4}$ & $10^{-4}$ \\
        Adam betas & $(0.9,0.99)$ & $(0.9,0.99)$ \\
        Adam epsilon & $10^{-6}$ & $10^{-6}$ \\
        Weight decay & $10^{-2}$ & $10^{-2}$ \\
        LR warm-up updates & 1 & 100 \\
        Precision & FP16 & FP16 \\
        Micro-batch size & 1 & 1 \\
        Data-loader workers & 8 & 4 \\
        Prefetch factor & 4 & 2 \\
        \bottomrule
    \end{tabular}
\end{table}


\paragraph{Sampling.}
At inference time, normal features and image-condition features
remain fixed while the texture latent evolves from Gaussian noise
to the predicted appearance representation.
Classifier-free guidance combines the image-conditioned and
image-dropped velocity predictions:
\begin{equation}
    u_{\mathrm{cfg}}
    =
    u_{\mathrm{drop}}
    +
    s\left(
        u_{\mathrm{cond}}-u_{\mathrm{drop}}
    \right),
    \label{eq:app_cfg}
\end{equation}
where $s$ is the guidance scale and both branches retain the
target geometry condition.

We use Euler integration over the noise-to-data interval $[0,1]$.
The configured sampler constructs 50 uniformly spaced time points,
corresponding to 49 Euler updates.
The final texture latent is converted to the VAE's scaled
representation and decoded into RGB values at the target
voxel locations.

\section{Experiment Settings}
\label{app:setting}

For the VAE, we primarily compare Tex-Zero VAE with FLUX VAE, TRELLIS.2, NaTex, and our model trained on 3D data. For FLUX VAE, TRELLIS.2, and NaTex, we use their official checkpoints and default settings. The compression settings of VAE for FLUX and TRELLIS.2 are f8c16 and f16c32, respectively. The VAE for Natex is a point-query vecset autoencoder that compresses surface points into latent tokens of 64 channels (80× point-to-token downsampling), producing an unstructured latent rather than a spatially downsampled feature grid.  For the 3D-trained model, we train the same VAE architecture on approximately 1M high-quality 3D assets from our internal dataset, while keeping other training settings unchanged.
For 3D reconstruction, we evaluate all models on an internal test set of 200 real 3D assets with rich and fine-grained appearance details. For 2D image reconstruction, we follow prior work and evaluate on the validation set of ImageNet.

For the DiT, we primarily compare Tex-Zero with TRELLIS and NaTex. In practical 3D generation pipelines, a user-provided image is first converted into multiple views using novel-view synthesis methods, which are then used to reconstruct the 3D geometry \cite{hunyuan3d}. The resulting geometry and multi-view images are subsequently used to generate the 3D texture. Tex-Zero focuses on this final stage, i.e., generating 3D textures conditioned on multi-view images and the corresponding 3D geometry. We therefore evaluate the multi-view image-to-3D texture generation capability of different methods.
NaTex supports multi-view image inputs, whereas the publicly released TRELLIS checkpoint supports only a single input view. To enable a fair comparison across methods, we focus on the appearance information provided by the input views and evaluate how faithfully each method transfers this information onto the target 3D geometry. We conduct the evaluation on an internal test set of 160 samples, each consisting of a 3D shape and its corresponding multi-view images. We report the differences between the multi-view renderings of the generated textured 3D assets and the corresponding ground-truth images.

\end{document}